\documentclass{article}
\PassOptionsToPackage{table}{xcolor}
\ifdefined\XeTeXversion
  \usepackage{iclr2027_conference}
  \usepackage{fontspec}
  \IfFileExists{/usr/share/fonts/opentype/urw-base35/NimbusRoman-Regular.otf}{%
    \setmainfont{NimbusRoman-Regular.otf}[Path=/usr/share/fonts/opentype/urw-base35/,BoldFont=NimbusRoman-Bold.otf,ItalicFont=NimbusRoman-Italic.otf,BoldItalicFont=NimbusRoman-BoldItalic.otf]
    \setsansfont{NimbusSans-Regular.otf}[Path=/usr/share/fonts/opentype/urw-base35/,BoldFont=NimbusSans-Bold.otf,ItalicFont=NimbusSans-Italic.otf,BoldItalicFont=NimbusSans-BoldItalic.otf]
    \setmonofont{NimbusMonoPS-Regular.otf}[Path=/usr/share/fonts/opentype/urw-base35/,BoldFont=NimbusMonoPS-Bold.otf,ItalicFont=NimbusMonoPS-Italic.otf,BoldItalicFont=NimbusMonoPS-BoldItalic.otf]
  }{}
\else
  \usepackage{iclr2027_conference,times}
\fi
\iclrfinalcopy
\usepackage{hyperref}
\hypersetup{hidelinks,pdfauthor={Xu Wang; Xunkai Li; Yinlin Zhu; Rong-Hua Li},pdftitle={GraphSelect for Budgeted Representation Selection in Multimodal Graph Inference},hypertexnames=false}
\usepackage{url}
\usepackage{graphicx}
\usepackage{amsmath,amssymb}
\usepackage{booktabs}
\usepackage{array}
\usepackage{multirow}
\usepackage[table]{xcolor}
\usepackage{placeins}
\usepackage{float}
\graphicspath{{figures/}}
\newcommand{\method}{\textsc{GraphSelect}}
\newcommand{\greedy}{\textsc{Forward}}

\newcommand{\cE}{\mathcal{E}}
\newcommand{\cF}{\mathcal{F}}

\newcommand{\cT}{\mathcal{T}}

\newcommand{\TV}{\operatorname{TV}}
\newcommand{\1}{\mathbf{1}}
\ifdefined\GraphSelectTableMacrosLoaded
\else
\def\GraphSelectTableMacrosLoaded{}

\newcommand{\tabhead}[1]{\textbf{#1}}

\newcommand{\tabstat}[2]{#1{\tiny\,$\pm$\,#2}}
\newcommand{\tabbeststat}[2]{\textbf{#1}{\tiny\,$\pm$\,#2}}
\newcommand{\tabsecondstat}[2]{\underline{#1}{\tiny\,$\pm$\,#2}}
\newcommand{\tabmainstat}[2]{#1{\tiny\,$\pm$#2}}
\newcommand{\tabmainbeststat}[2]{\textbf{#1}{\tiny\,$\pm$#2}}
\newcommand{\tabmainsecondstat}[2]{\underline{#1}{\tiny\,$\pm$#2}}

\newcommand{\graphselectMainTableSetup}{%
  \footnotesize\setlength{\tabcolsep}{1.2pt}\renewcommand{\arraystretch}{1.05}}

\newcommand{\graphselectCompactTableSetup}{%
  \footnotesize\setlength{\tabcolsep}{2.4pt}\renewcommand{\arraystretch}{1.05}}
\newcommand{\graphselectSearchTableSetup}{%
  \small\setlength{\tabcolsep}{1.8pt}\renewcommand{\arraystretch}{1.05}}

\fi
\title{GraphSelect for Budgeted Representation Selection in Multimodal Graph Inference}
\author{
Xu Wang$^{1}$ \quad Xunkai Li$^{2}$ \quad Yinlin Zhu$^{3}$ \quad Rong-Hua Li$^{2}$\\[8pt]
$^{1}$School of Airspace Science and Engineering, Shandong University, Weihai, China\\
$^{2}$Department of Computer Science, Beijing Institute of Technology, Beijing, China\\
$^{3}$School of Computer Science and Engineering, Sun Yat-sen University, Guangzhou, China\\[4pt]
\textbf{Correspondence to:} Rong-Hua Li \texttt{<lironghuabit@126.com>}
}

\begin{document}
\maketitle
\raggedbottom
\begin{abstract}
Multimodal graph predictors combine text, images, and relations to classify connected entities. How much of this input is needed to preserve their predictions? We study budgeted representation selection, which chooses a subset of candidate text and image vectors under a separate capacity for each modality. Predictions from the complete candidate input define the classes to preserve. The challenge is that a representation's contribution depends on the other selected inputs, while graph propagation extends its effects across nodes. Our empirical study shows that candidate rankings change with the selected input, while predicted probabilities remain informative after the class stops changing. Updating scores improves selection, and exchanging inputs can improve a subset whose capacity is already filled. These findings lead to \method{}, which starts from individual candidate gains and refines the subset through jointly evaluated exchanges. It screens promising removals and additions, accepts an exchange when it reduces the prediction loss, and updates the scores. Experiments on six graphs show higher mean objective recovery than six attribution and explanation methods adapted to the selection task. Across nine trained architectures on two graphs, retaining 20\% of the candidate representations per modality gives a mean accuracy drop of 0.10 percentage points relative to full candidate input, preserving classification performance with substantially fewer text and image representations.
\end{abstract}

\section{Introduction}

Multimodal attributed graphs associate connected entities, such as products, artworks, and social media posts, with text and images~\citep{MultimodalGraphs,MultimodalGraphSurvey,MAGB,OpenMAG}. By combining these attributes with relations, graph models support recommendation~\citep{MMGCN,MGAT,LGMRec}, cross-modal matching~\citep{GSMN}, and node analysis~\citep{mmgraph,DMGC,DGF,CoMAG}. We ask which encoded text and image representations should remain available to preserve a trained graph predictor's decisions under limited input capacity.

We formulate this problem as \emph{budgeted representation selection for multimodal graph inference}. Each modality has a prescribed number of input slots. We choose which candidate representations occupy those slots and evaluate the resulting input with the trained predictor. Selection has access to all candidates, and the full-input predicted classes provide its targets. The graph, predictor, and representations outside the candidate pool remain unchanged. Unselected candidates use their modality's mean training representation. Ground-truth labels are not used during selection.

\begin{figure}[t]
\centering
\includegraphics[width=\linewidth]{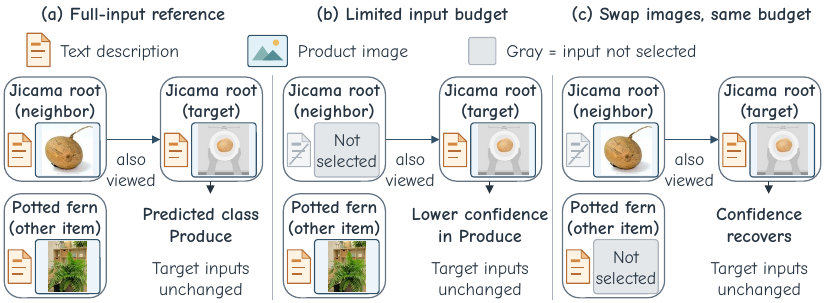}
\caption{\textbf{A neighboring image can help preserve a graph prediction.} Each box represents a product. The two jicama products have an also-viewed relation. Full input provides the reference prediction. With a limited budget, omitting the neighbor's image lowers confidence in Produce. Exchanging the fern's image for the neighbor's restores most of that confidence at the same budget. The target's own inputs remain available. Gray inputs are not selected and use modality means.}
\label{fig:task-setting}
\end{figure}

Selection matters beyond a candidate's own node. In the Grocery example in Fig.~\ref{fig:task-setting}, a target product keeps its text and image, yet loses support for its predicted category when a neighboring image is unavailable. Exchanging another selected image for that neighbor's image restores most of this support at the same capacity. Scoring the two images individually favors the fern image, but evaluating their role in the selected subset favors this exchange.

Our empirical study identifies three decisions behind this difference. Candidate scores should reflect the input being assembled, reward the original predicted classes, and change as the subset changes. Revising earlier choices then improves the completed selection.

These findings motivate \method{}, which refines an initial subset by jointly evaluating promising exchanges and updating scores after an improvement. It can thereby replace earlier choices that have become less useful. We compare it with \greedy{}, which updates addition gains and keeps every earlier selection. For additive graph predictors, \method{} evaluates changes exactly on the affected target nodes.

\noindent\textbf{Our Contributions.}
(1) \textbf{In-depth Investigation.} We formulate representation selection under separate modality capacities and examine how the scoring input, prediction objective, and score updates affect selection quality in multimodal graphs.
(2) \textbf{Novel Method.} We introduce \method{}, which refines an initial subset through screened exchanges. Joint evaluation accounts for interactions within each replacement, while score updates adapt the search to the selected subset.
(3) \textbf{Empirical Evaluation.} Across six graphs, \method{} outperforms six attribution and explanation methods adapted to the selection task. Experiments with nine trained architectures on two graphs show how increasing capacity improves objective recovery and reduces downstream prediction gaps.

\FloatBarrier
\section{Related Works}
\label{sec:related}

\textbf{Multimodal graph learning.}
Multimodal attributed graphs combine relations with heterogeneous node attributes~\citep{MultimodalGraphs,MultimodalGraphSurvey,MAGB,OpenMAG,mmgraph}. Graph convolution and attention integrate these inputs for recommendation~\citep{MMGCN,MGAT,LGMRec}, while semantic matching, filtering, and dynamic graph construction capture interactions across modalities and nodes~\citep{GSMN,DMGC,DGF,CoMAG}. Recent models improve how this context is used. LION couples modality alignment and fusion through graph propagation~\citep{LION}, and RoleMAG assigns neighbors different propagation roles according to their contribution~\citep{RoleMAG}. Other directions address generative tasks and cold-start nodes~\citep{MMGL,NTSFormer}, global transformers~\citep{MIGGT}, and unified embeddings or language model integration~\citep{Graph4MM,GraphMLLM,GraphGPTO,MLaGA,UniGraph2}. GraphSelect studies how these trained predictors preserve their decisions when a subset of encoded node inputs is retained.

\textbf{Modality selection and acquisition.}
Modality selection studies which information sources provide complementary predictive value. Greedy modality selection formalizes this problem through a utility function with approximate submodularity~\citep{GreedyModalitySelection}. CAMA allocates the acquisition of an additional modality across a cohort by estimating the utility of the missing information~\citep{CAMA}. Data selection instead chooses examples that contribute to model training, using quantities such as gradient matching or data value~\citep{GradMatch,DataShapleySelection}. Our decision unit is one node's encoded text or image representation. Selection allocates separate capacities across these available representations and evaluates their joint effect on graph predictions. 

\textbf{Attribution and graph explanation.}
Feature attribution assigns contributions to model inputs through gradients, perturbations, or Shapley values~\citep{IntegratedGradients,Captum,Shapley}. GNNExplainer identifies a compact subgraph and feature subset for a prediction~\citep{GNNExplainer}, while SubgraphX, GNNShap, and GOAt develop graph-specific contribution estimates~\citep{SubgraphX,GNNShap,GOAt}. GraphFramEx emphasizes evaluating explanations according to their sufficiency and necessity for the prediction~\citep{GraphFramEx}. Their scores provide a starting point for selecting a common subset across nodes and modalities. Our empirical study examines how the scored input, objective, and score updates affect this decision.

\textbf{Prediction-preserving input subsets.}
Sufficient input subsets identify small sets of observed features that retain a model decision, using backward selection~\citep{SufficientInputSubsets}. Zorro constructs sparse graph explanations whose predictions remain stable when omitted inputs are randomly perturbed~\citep{Zorro}. Our setting evaluates a common subset of node-level representations over a target-node set, with a separate capacity for each modality. GraphSelect optimizes probability assigned to the full-input classes and revises the subset through jointly evaluated exchanges. Zorro is included in the comparison with its randomized class-retention objective, allowing the two selection criteria to be examined within the same candidate and capacity setup.

\section{Problem Formulation and Empirical Study}
\subsection{Problem Formulation}
\label{sec:problem}

\textbf{Graph, predictor, and candidate pool.}
Let $\mathcal G=(\mathcal V,\mathcal E_G)$ be a multimodal attributed graph with adjacency matrix $A$, and let $H$ contain its encoded text and image representations. A trained predictor $f_\theta(A,H)$ produces class logits at each node. Selection changes only the availability of candidate representations, leaving the graph, encodings, and predictor parameters unchanged. We designate $N$ candidate nodes whose text and image vectors are subject to capacity limits, and a target set $\cT$ on which predictions are evaluated. Candidate and target nodes may overlap. Representations outside the candidate pool provide context throughout selection.

\textbf{Selectable representation and reference input.}
Each candidate node contributes one text vector and one image vector. We refer to either vector, together with its node identity, as a candidate representation. With modality set $\mathcal M=\{\mathrm{text},\mathrm{image}\}$, the candidates form
\begin{equation}
\cE=\mathop{\dot\bigcup}_{m\in\mathcal M}\cE_m,
\qquad |\cE_m|=N\quad\text{for every }m\in\mathcal M.
\end{equation}
Candidate $e$ has an observed representation $h_e$ and a reference $r_e$, defined as the mean representation over training nodes for the same modality. A selected set $S\subseteq\cE$ determines the predictor input
\begin{equation}
H_S(e)=
\begin{cases}
h_e,&e\in S,\\
r_e,&e\in\cE\setminus S.
\end{cases}
\label{eq:selected-state}
\end{equation}
where noncandidate entries of $H_S$ keep their observed values. Write $Z_S=f_\theta(A,H_S)$ and $P_S=\operatorname{softmax}(Z_S)$. The reference input $S=\varnothing$ replaces all candidates by their modality means, whereas full candidate input $S=\cE$ uses all observed representations. Before selection, we run the predictor on full candidate input to obtain the target classes.

\textbf{Output and objective.}
Let $P_{\cE}$ be the full-input probabilities and $\widehat y_t=\operatorname*{arg\,max}_c P_{\cE,tc}$ the predicted class for target $t\in\cT$. We measure how well a selected subset supports these classes by
\begin{equation}
R(S)=-\frac{1}{|\cT|}\sum_{t\in\cT}\log P_{S,t\widehat y_t}.
\label{eq:full-input-class-ce}
\end{equation}
We seek a joint subset with low $R(S)$. The loss remains sensitive to probability changes even when the predicted class stays the same. A subset may assign that class greater probability than full input, so $R(S)$ is not a distance to $P_{\cE}$. Sec.~\ref{sec:empirical} compares it with matching the complete probability distribution.

\textbf{Separate modality capacities.}
Let $\mathbf K=(K_{\mathrm{text}},K_{\mathrm{image}})$ denote the available text and image budgets, with $0\le K_m\le |\cE_m|$. Each selected representation consumes one unit of its modality budget. The feasible family and selection objective are
\begin{equation}
S^\star\in\operatorname*{arg\,min}_{S\in\cF_{\mathbf K}}R(S),
\qquad
\cF_{\mathbf K}=\{S\subseteq\cE\mid |S\cap\cE_m|=K_m,\ \forall m\in\mathcal M\}.
\label{eq:selection-objective}
\end{equation}
The two quotas specify the number of representations retained from each modality.

\subsection{Empirical Study}
\label{sec:empirical}
\FloatBarrier

The selection objective requires evaluating the input assembled from the retained representations. We examine three choices that determine whether candidate scores guide this process. The first is where to evaluate a candidate, by deletion from full input or addition to the reference input. The second is whether to optimize predicted-class probability, hard agreement, or the full probability distribution. The third is whether to update scores as the subset changes.

\textbf{Study design.}
We vary one choice at a time while using the same graph, predictor, candidates, targets, modality references, budgets, and tie rule. The deletion probe covers five multimodal graphs~\citep{MAGB,OpenMAG}, three split seeds, and 220 candidate nodes per setting. Cross-entropy deletion (CE deletion) replaces one candidate by its modality mean at full input and measures the increase in Eq.~\eqref{eq:full-input-class-ce}. Total variation deletion (TV deletion) measures half the sum of absolute probability differences over classes, averaged over target nodes. We compare these scores with exact first-addition gains for 6,600 candidate representations.

The KL divergence control reduces Kullback-Leibler divergence from the full-input probabilities. Prediction agreement maximizes the fraction of target nodes retaining their full-input predicted class. The initial-gain ranking computes each candidate's loss reduction once at the reference input. \greedy{} instead recomputes gains after every addition. These comparisons add EleFashion~\citep{mmgraph}, giving six graphs and 54 combinations of graph, seed, and budget. We express objective recovery relative to the loss reduction from reference to full input, with 0\% and 100\% marking these two losses. Tables report the mean recovery across settings. The pooled recovery reported below is computed after pooling losses over affected target nodes, so it can differ from the mean of the individual ratios.

\begin{figure}[t]
\centering
\includegraphics[width=0.98\linewidth]{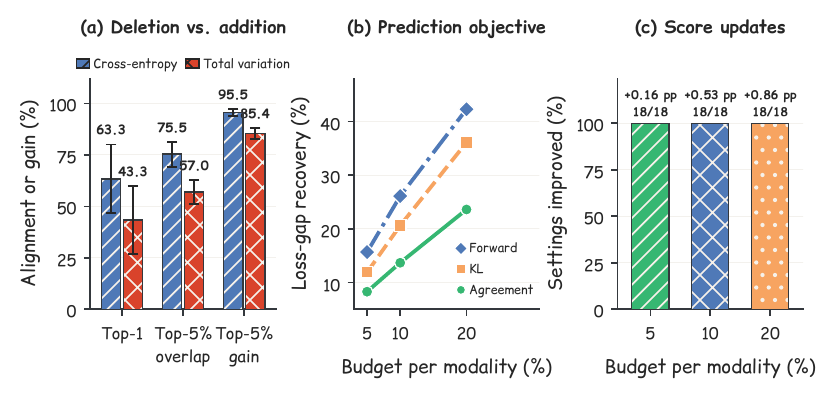}
\caption{\textbf{Three design probes.} (a) Agreement between deletion rankings and first-addition gains. (b) Recovery under three prediction objectives. (c) Improvements from updating addition scores at each step.}
\label{fig:empirical-study}
\end{figure}

\textbf{Deletion and addition favor different candidates.}
CE deletion identifies the best first addition in 19 of 30 graph, seed, and modality groups, compared with 13 for TV deletion (Fig.~\ref{fig:empirical-study}a). Their mean Spearman correlations with addition gains are 0.950 and 0.791. CE deletion therefore provides a stronger initial ranking, yet misses the best addition in 11 groups. The distinction is the surrounding input. Deletion measures what a representation contributes alongside all other candidates, whereas construction asks what it contributes to the subset being assembled. The Appendix gives further ranking comparisons and a counterexample.

\textbf{Predicted-class probability remains informative after hard agreement saturates.}
At 5\%, 10\%, and 20\% capacity, predicted-class cross-entropy achieves pooled recovery of 15.7\%, 26.1\%, and 42.3\%. KL divergence reaches 11.9\%, 20.6\%, and 36.1\%, while prediction agreement reaches 8.3\%, 13.7\%, and 23.6\%. The difficulty with hard agreement is visible during construction. It gives zero gain at 1,877 of 2,772 selected steps, compared with none for predicted-class cross-entropy. Agreement is already near saturation, averaging 99.85\% for \greedy{} and 99.95\% for the agreement control over affected targets and budgets. Cross-entropy continues to reward additional probability assigned to the target class. KL divergence also remains smooth, but requires matching probabilities for the other classes. These results favor predicted-class cross-entropy for the selection task.

\textbf{Updating scores improves selection consistently.}
Updating gains lowers the loss in all 54 comparisons with the initial-gain ranking (Fig.~\ref*{fig:empirical-study}c). The mean recovery improvement grows from 0.2 percentage points at 5\% capacity to 0.9 at 20\%. Pooled recovery is 15.7\%, 26.1\%, and 42.3\% for \greedy{}, compared with 15.5\%, 25.6\%, and 41.4\% for the initial-gain ranking and 15.2\%, 25.1\%, and 41.2\% for CE deletion. These consistent reductions show that an initial ranking leaves useful improvements available as the subset grows.

\textbf{Role of the graph.}
Removing feature and label propagation reduces mean full-input accuracy from 76.38\% to 70.41\%, with lower accuracy on five of six datasets. Propagation also expands a candidate representation's effect from its own node to an average of 127.2 target nodes. The graph thus supplies predictive context and determines which predictions an input change can affect. Appendix~\ref{sec:graph-ablation} reports the controlled comparisons.

\textbf{Design implication.}
The probes favor an objective that rewards probability assigned to the full-input classes and scores that reflect the current subset. An exchange extends score updates to revising earlier choices, replacing a representation when another becomes more useful within the current subset. The exchange controls find replacements that improve some completed \greedy{} subsets while preserving both quotas. This motivates refining the selected set through exchanges, as developed in Sec.~\ref{sec:method}.

\section{Methodology}
\label{sec:method}

\subsection{Overview}

\method{} fills both modality quotas from an initial ranking, then revises the subset through screened exchanges. The predictor evaluates each proposed exchange jointly over the target nodes, and an accepted improvement triggers new scores. Sec.~\ref{sec:sparse} shows how additive graph predictors restrict these calculations to the affected nodes.

\begin{figure*}[t]
  \centering
  \includegraphics[width=0.98\textwidth]{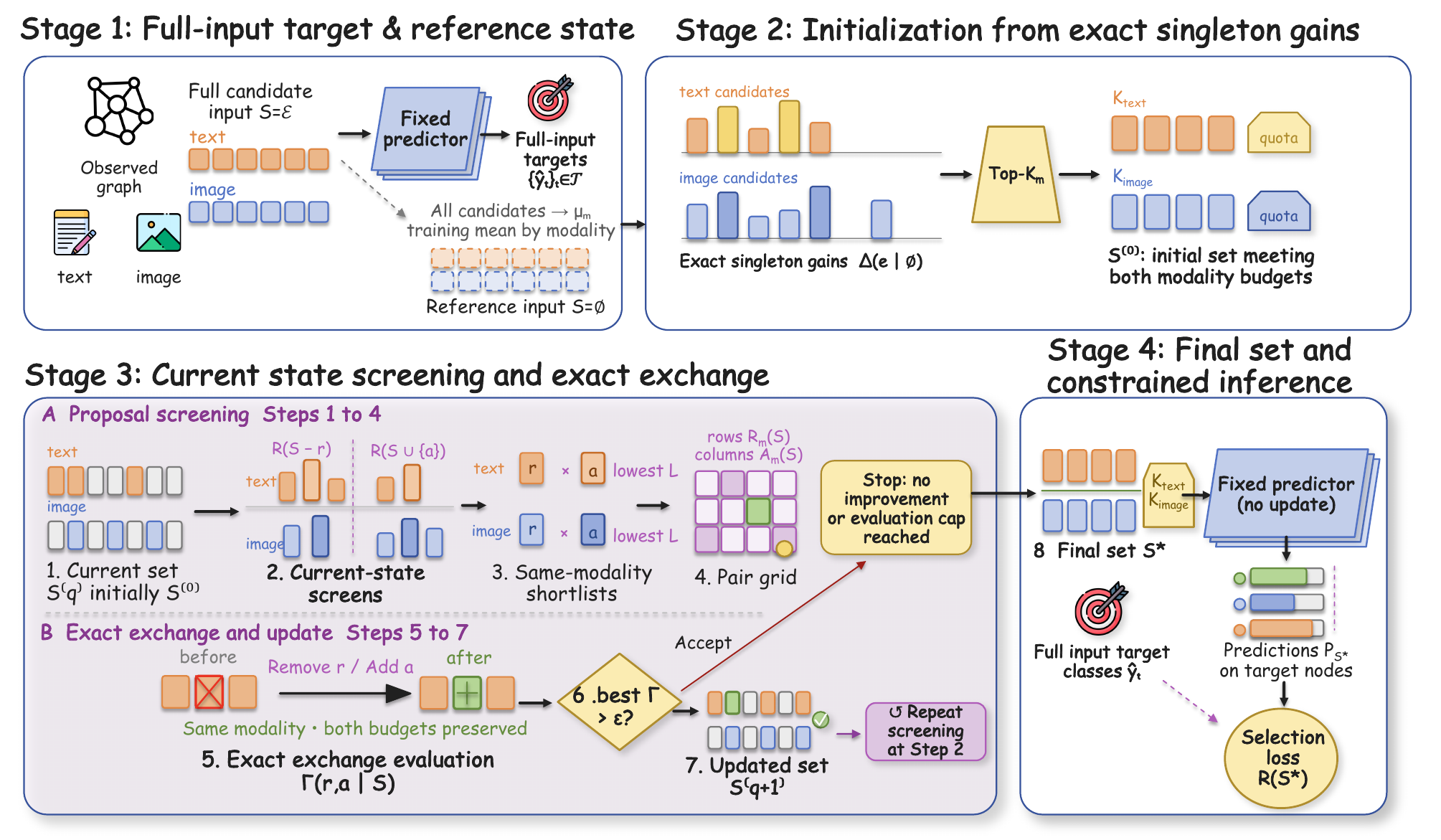}
  \caption{\method{} overview. The graph predictor supplies the full-input target classes and evaluates each proposed subset. Singleton gains initialize selection, followed by screening and joint exchange evaluation. Accepted exchanges preserve both modality quotas and trigger score updates.}
  \label{fig:framework}
\end{figure*}

\subsection{Initialization and Current State Screening}

Initialization provides a feasible subset from which to search. The gain from adding candidate $e$ is $\Delta(e\mid S)=R(S)-R(S\cup\{e\})$. We evaluate each candidate at the reference input and retain the largest gains within each modality,
\begin{equation}
S^{(0)}=\bigcup_{m\in\mathcal M}\operatorname{TopK}_{K_m}
\{\Delta(e\mid\varnothing)\mid e\in\cE_m\}.
\label{eq:addition-initialization}
\end{equation}
Here $\operatorname{TopK}_{K_m}$ returns the $K_m$ candidate identities with the highest scores, resolving ties by a predetermined order. This initialization fills both quotas with one score per candidate.

The search then identifies promising replacements within the current subset. For modality $m$, the removal shortlist $\mathcal R_m(S)$ contains the $L$ selected candidates with the lowest removal losses $R(S\setminus\{r\})$. The addition shortlist $\mathcal A_m(S)$ contains the $L$ unselected candidates with the lowest addition losses $R(S\cup\{a\})$. Each list uses all available candidates if fewer than $L$ remain. Removal scores identify inputs the subset can most readily give up, while addition scores identify inputs that improve its present predictions. These scores screen candidates for joint evaluation.

\subsection{Exact Screened Exchange}

A useful addition can overlap with the contribution of the input being removed. GraphSelect therefore evaluates the two changes together. The candidate pairs form $\mathcal P(S)=\bigcup_{m\in\mathcal M}\mathcal R_m(S)\times\mathcal A_m(S)$, so each pair replaces one representation by another of the same modality. At iteration $q$, we choose
\begin{equation}
(r_q,a_q)\in\operatorname*{arg\,max}_{(r,a)\in\mathcal P(S^{(q)})}
\left[R(S^{(q)})-R\bigl((S^{(q)}\setminus\{r\})\cup\{a\}\bigr)\right].
\label{eq:screened-exchange}
\end{equation}
Denote the bracketed loss reduction by $\Gamma(r,a\mid S^{(q)})$. If the best reduction exceeds $\varepsilon$, we update $S^{(q+1)}=(S^{(q)}\setminus\{r_q\})\cup\{a_q\}$ and recompute both shortlists. Otherwise, the search terminates. Recomputing the screens lets the next exchange reflect the inputs just retained, rather than continue from the initial ranking.

Each accepted exchange preserves both quotas and strictly lowers the loss. GraphSelect thus returns a feasible subset whose loss is no greater than its initial loss. We use $\varepsilon=10^{-12}$ to reject numerical ties. In the budgeted comparison, the shortlist width is $L\le8$ and the total allowance is $\lfloor0.35Q_{\mathrm{forward}}\rfloor$ objective evaluations, including initialization, where $Q_{\mathrm{forward}}$ is the count needed by Forward to fill both quotas. The search reduces $L$ as this allowance is consumed and stops when it cannot evaluate another exchange. The Appendix gives the algorithm and proof.

\subsection{Exact Evaluation for Additive Predictors}
\label{sec:sparse}

The controlled benchmark uses an additive graph predictor, allowing candidate changes to be evaluated on affected targets. Let $x_i^m$ be the log-probability vector for node $i$ and modality $m$, and write $U_i=\frac12\sum_{m\in\mathcal M}x_i^m$. With row-normalized adjacency $\widetilde A$, smoothed label propagation probabilities $Q$, term weights $\lambda$, and hop weights $\eta$, the logits are
\begin{equation}
Z=\lambda_0U+\lambda_g(\eta_1\widetilde AU+\eta_2\widetilde A^2U)
+\lambda_{\mathrm{LP}}\log Q.
\label{eq:additive-predictor}
\end{equation}
Let $B_e$ be the logit change when candidate $e$ replaces its reference value. The selected input gives $Z_S=Z_\varnothing+\sum_{e\in S}B_e$. Although logits add, cross-entropy couples their contributions, so the value of a candidate depends on the other retained inputs. Graph propagation determines the target rows that each $B_e$ can affect, extending to two hops in this predictor.

Both additions and exchanges can be evaluated through one logit change $D$. Let $\cT_D$ contain its affected target rows and $\mathcal L_t(z)=-\log\operatorname{softmax}(z)_{\widehat y_t}$. The loss reduction is
\begin{equation}
g(D\mid S)=\frac{1}{|\cT|}\sum_{t\in\cT_D}
\left[\mathcal L_t(Z_{S,t})-\mathcal L_t(Z_{S,t}+D_t)\right].
\label{eq:sparse-transitions}
\end{equation}
An addition uses $D=B_e$, giving $\Delta(e\mid S)=g(B_e\mid S)$, while an exchange uses $D=B_a-B_r$, giving $\Gamma(r,a\mid S)=g(B_a-B_r\mid S)$. Loss terms outside the affected rows cancel exactly. Indexing these rows therefore preserves the objective while avoiding recomputation over unchanged targets. Nonlinear predictors evaluate the same subset changes through complete forward passes. The Appendix describes gain reuse and evaluation counts.

\section{Experiments}
\label{sec:experiments}
\raggedbottom

We compare \method{} with attribution and explanation methods, then examine its components and the effect of representation capacity.

\subsection{Experimental Settings}

\textbf{Datasets and protocol.}
The additive comparison uses six multimodal graph benchmarks~\citep{MAGB,OpenMAG,mmgraph}, three split seeds, $N=220$ candidate nodes, and capacities $b$ of 5\%, 10\%, and 20\%, with $K_m=\lceil bN\rceil$. We also evaluate nine trained nonlinear architectures on Grocery and RedditS at 2.5\%, 5\%, 10\%, 15\%, 20\%, and 30\%. Test-set membership and test labels are unavailable to the selectors and used only for downstream evaluation.

\textbf{Comparators.}
We adapt the six published methods in Table~\ref{tab:main-external} to score the same candidate representations~\citep{Shapley,Captum,IntegratedGradients,GNNExplainer,GOAt,Zorro}. Each receives the same predictor, target nodes, and modality quotas. Sec.~\ref{sec:empirical} defines the additional controls, while the Appendix explains the adaptation of each published method.

\textbf{Metrics.}
We report selection loss $R(S)$ and objective recovery. Prediction agreement, total variation, accuracy, Macro-F1, and cross-entropy to ground-truth labels describe how selection changes the predictions. Tables report means and standard deviations over the stated settings. Hierarchical uncertainty intervals resample datasets and then seeds.

\FloatBarrier
\subsection{Comparison Under the Shared Protocol}

\begin{figure}[t]
  \centering
  \includegraphics[width=0.96\linewidth]{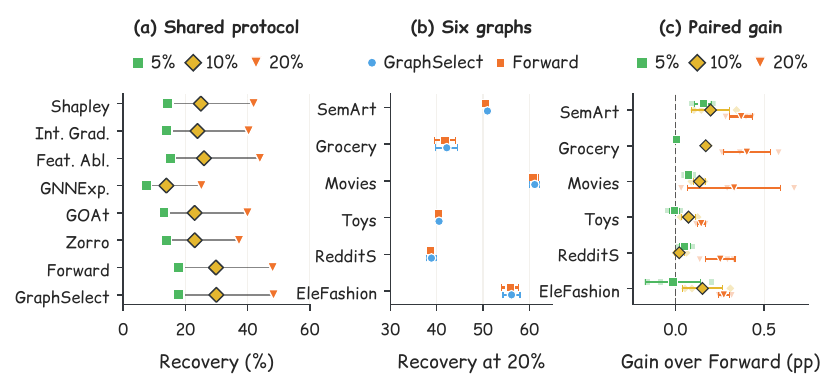}
  \caption{\textbf{Selection quality.} (a) Mean recovery at three capacities. (b) Dataset results at 20\% capacity. (c) Paired recovery gains over Forward. All methods use the same candidate pool and modality quotas.}
  \label{fig:shared-budget-overview}
\end{figure}

Table~\ref{tab:main-external} compares objective recovery across the six graphs. Figure~\ref*{fig:shared-budget-overview} shows the corresponding budget trends and differences from \greedy{}.

\begin{table}[H]
\caption{\textbf{Objective recovery (\%).} Dataset columns summarize three seeds, and overall means summarize 18 graph and seed settings. Values are mean $\pm$ standard deviation. GraphSelect uses the budgeted search setting.}
\label{tab:main-external}
\centering
\graphselectMainTableSetup
\begin{tabular}{lrrrrrrrrr}
\toprule
\multirow{2}{*}{\tabhead{Method}} & \multicolumn{6}{c}{\tabhead{Recovery at 20\% capacity}} & \multicolumn{3}{c}{\tabhead{Mean over 18 settings}} \\
\cmidrule(lr){2-7}\cmidrule(lr){8-10}
 & \tabhead{SemArt} & \tabhead{Grocery} & \tabhead{Movies} & \tabhead{Toys} & \tabhead{RedditS} & \tabhead{EleFashion} & \tabhead{5\%} & \tabhead{10\%} & \tabhead{20\%} \\
\midrule
\multicolumn{10}{l}{\textbf{Attribution and explanation adapters}} \\
Shapley Sampling & \tabmainstat{44.7}{1.2} & \tabmainstat{37.3}{1.5} & \tabmainstat{53.7}{1.1} & \tabmainstat{36.1}{0.2} & \tabmainstat{32.1}{1.3} & \tabmainstat{47.5}{4.2} & \tabmainstat{14.3}{3.4} & \tabmainstat{24.9}{5.1} & \tabmainstat{41.9}{7.7} \\
Integrated Gradients & \tabmainstat{44.2}{2.3} & \tabmainstat{37.1}{1.1} & \tabmainstat{51.7}{1.5} & \tabmainstat{35.2}{0.1} & \tabmainstat{30.2}{0.7} & \tabmainstat{43.3}{4.1} & \tabmainstat{13.9}{3.7} & \tabmainstat{23.9}{5.2} & \tabmainstat{40.3}{7.3} \\
Feature Ablation & \tabmainsecondstat{47.5}{0.7} & \tabmainsecondstat{37.7}{1.7} & \tabmainstat{55.2}{1.9} & \tabmainsecondstat{37.0}{0.2} & \tabmainsecondstat{33.7}{1.2} & \tabmainsecondstat{52.5}{2.6} & \tabmainsecondstat{15.2}{3.7} & \tabmainsecondstat{26.0}{5.4} & \tabmainsecondstat{43.9}{8.3} \\
GNNExplainer & \tabmainstat{22.5}{1.2} & \tabmainstat{28.0}{1.6} & \tabmainstat{29.0}{1.8} & \tabmainstat{24.9}{0.8} & \tabmainstat{18.0}{0.4} & \tabmainstat{28.3}{3.9} & \tabmainstat{7.3}{2.3} & \tabmainstat{13.9}{3.5} & \tabmainstat{25.1}{4.4} \\
GOAt & \tabmainstat{39.7}{2.0} & \tabmainstat{37.6}{1.6} & \tabmainsecondstat{57.4}{1.9} & \tabmainstat{36.7}{0.3} & \tabmainstat{29.8}{1.3} & \tabmainstat{38.3}{4.9} & \tabmainstat{13.1}{4.2} & \tabmainstat{22.9}{6.2} & \tabmainstat{39.9}{8.8} \\
\midrule
\multicolumn{10}{l}{\textbf{Sufficient-subset adapter}} \\
Zorro & \tabmainstat{43.7}{1.5} & \tabmainstat{31.2}{1.8} & \tabmainstat{44.1}{3.2} & \tabmainstat{28.6}{0.7} & \tabmainstat{31.5}{0.3} & \tabmainstat{44.2}{2.6} & \tabmainstat{13.9}{3.7} & \tabmainstat{23.0}{5.5} & \tabmainstat{37.2}{7.1} \\
\midrule
\multicolumn{10}{l}{\textbf{Proposed method}} \\
\textbf{\method{}} & \tabmainbeststat{50.9}{0.5} & \tabmainbeststat{42.2}{2.3} & \tabmainbeststat{61.1}{1.0} & \tabmainbeststat{40.5}{0.4} & \tabmainbeststat{38.9}{1.0} & \tabmainbeststat{56.1}{1.8} & \tabmainbeststat{17.9}{4.0} & \tabmainbeststat{29.9}{6.1} & \tabmainbeststat{48.3}{8.5} \\
\bottomrule
\end{tabular}
\end{table}

\textbf{GraphSelect achieves the highest mean recovery.}
At 5\%, 10\%, and 20\% capacity, \method{} recovers 17.9\%, 29.9\%, and 48.3\% of the objective gap on average. Feature Ablation, the strongest published comparator by mean recovery, reaches 15.2\%, 26.0\%, and 43.9\%. The advantage holds across the three capacities and all six dataset means at 20\%, showing that joint refinement improves on selecting candidates solely from attribution or explanation scores.

\textbf{GraphSelect improves on the study controls.}
Compared with \greedy{}, budgeted \method{} gives lower loss in 46 of 54 settings, ties in five, and higher loss in three. Their mean recoveries reach 48.3\% and 48.0\% at 20\% capacity. Refining the completed subset extends the improvements obtained by repeated addition rescoring. The initial-gain ranking and CE deletion also have lower mean recovery than \method{}.

\subsection{Search Ablations and Capacity Analysis}
\label{sec:nonlinear}

We tune the shortlist width on a predefined subset of Grocery settings and use the selected width in 126 disjoint evaluation settings. Appendix Table~\ref{tab:search-ablation} reports the complete loss differences and win, tie, and loss counts.

We compare Forward with GraphSelect and its two ablations under matched evaluation budgets, treating loss differences within $10^{-6}$ as ties.

\textbf{Exchange refinement and score updates improve selection.}
Removing exchange refinement increases the loss in 112 of 126 settings and ties in 14. Keeping the initial scores during exchange increases the loss in 62 settings, ties in 62, and improves it in two. Against Forward, GraphSelect improves 57 settings, ties in 68, and loses in one. The component controls show that both revising the initial subset and recomputing the screens contribute to selection quality. The Forward comparison further shows that revising earlier choices improves on repeated addition within the same evaluation budget.

\textbf{Capacity and predictive performance.}
We vary the number of retained representations with the trained predictor and all other inputs unchanged. Accuracy and Macro-F1 show how selection affects test decisions, complementing probability recovery.

\begin{figure}[t]
  \centering
  \includegraphics[width=0.86\linewidth]{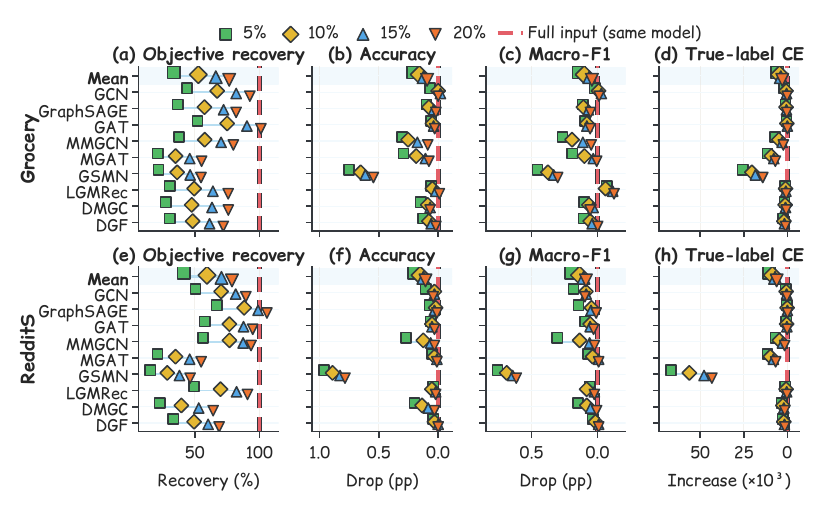}
  \caption{\textbf{Capacity and prediction quality.} Objective recovery and downstream performance gaps at four capacities on Grocery and RedditS. Red dashed lines indicate the corresponding full-input results.}
  \label{fig:budget-preservation}
\end{figure}

\textbf{Larger subsets reduce the prediction gap.}
Across 54 architecture, graph, and seed settings, mean recovery rises from 37.6\% at 5\% capacity to 77.6\% at 20\%. At 20\%, the mean accuracy and Macro-F1 drops are 0.100 and 0.065 percentage points. Recovery reaches 90.4\% at 30\%. Even at the smallest quota of six representations per modality, the mean accuracy gap is 0.237 percentage points. Table~\ref{tab:budget-sweep-exact} reports all six capacities, and Appendix Figure~\ref{fig:supp-selector-trajectories} compares the four selectors.

\begin{table}[H]
\centering
\caption{\textbf{GraphSelect across six capacities.} Values summarize 54 settings. Accuracy and Macro-F1 drops are in percentage points. The increase in cross-entropy to ground-truth labels is multiplied by $10^3$.}
\label{tab:budget-sweep-exact}
\graphselectCompactTableSetup
\begin{tabular}{@{}lrrrr@{}}
\toprule
\tabhead{Capacity (quota)} & \tabhead{Objective recovery (\%)} & \tabhead{Accuracy drop} & \tabhead{Macro-F1 drop} & \tabhead{CE increase} \\
\midrule
2.5\% (6) & \tabstat{24.81}{11.64} & \tabstat{0.237}{0.270} & \tabstat{0.202}{0.227} & \tabstat{9.14}{16.80} \\
5.0\% (11) & \tabstat{37.56}{15.50} & \tabstat{0.207}{0.256} & \tabstat{0.171}{0.201} & \tabstat{8.14}{15.62} \\
10.0\% (22) & \tabstat{56.02}{17.91} & \tabstat{0.156}{0.237} & \tabstat{0.119}{0.179} & \tabstat{6.53}{13.12} \\
15.0\% (33) & \tabstat{68.38}{18.27} & \tabstat{0.124}{0.225} & \tabstat{0.084}{0.172} & \tabstat{5.40}{11.31} \\
20.0\% (44) & \tabstat{77.61}{17.92} & \tabstat{0.100}{0.212} & \tabstat{0.065}{0.166} & \tabstat{4.52}{10.15} \\
30.0\% (66) & \tabsecondstat{90.39}{16.62} & \tabsecondstat{0.069}{0.176} & \tabsecondstat{0.040}{0.145} & \tabsecondstat{3.19}{7.50} \\
\midrule
\textbf{Full input} & \tabbeststat{100.00}{0.00} & \tabbeststat{0.000}{0.000} & \tabbeststat{0.000}{0.000} & \tabbeststat{0.00}{0.00} \\
\bottomrule
\end{tabular}
\end{table}

At 20\% capacity, the three graph architectures recover 91.0\% to 97.9\% of the objective gap, compared with 50.5\% to 86.6\% for the multimodal graph architectures. The mean accuracy gap is 0.667 percentage points for GSMN and 0.006 to 0.060 for the other eight architectures. These results separate two aspects of preservation. Changes in class probabilities need not change the most likely class, so a small accuracy gap can coexist with a larger objective gap. Increasing capacity improves probability recovery even when accuracy is already close to the full-input result. Appendix Table~\ref{tab:architecture-budget-detail} reports the individual architecture results.

\FloatBarrier

\section{Conclusion}

We presented GraphSelect for preserving multimodal graph predictions under separate text and image representation budgets. Its search follows the selected input, revising earlier choices through jointly evaluated exchanges and score updates. Graph propagation determines which target predictions a representation can support, and joint evaluation measures its contribution. GraphSelect achieves the highest mean objective recovery in the six-graph comparison. Across nine trained architectures, retaining 20\% of the candidate representations per modality yields a mean accuracy gap of 0.10 percentage points to full input. The results demonstrate how joint selection preserves predictive performance under a limited candidate representation budget.

\clearpage
\subsection*{AI use statement}
In this work, we used generative AI tools to aid and polish the writing of this manuscript, including grammar, phrasing, and clarity, as the authors are non-native English speakers. We have reviewed all AI-assisted work, and the authors verified the final text against the reported results and claims. We take responsibility for the final content of this work, including text, claims, or artifacts produced with the aid of generative AI.

\subsection*{Reproducibility statement}
The problem formulation, method, and appendix specify the graph operators, selection objective, candidate construction, reference representations, evaluation protocol, and the adaptations applied to external methods. The mathematical analysis states the assumptions behind the reported claims, and the appendix reports the complete selection results, ablations, and computational cost. Code will accompany the preprint.

{\small
\bibliographystyle{iclr2027_conference}
\bibliography{references}
}

\appendix
\raggedbottom
\renewcommand{\theHsection}{appendix.\Alph{section}}
\renewcommand{\theHfigure}{appendix.\arabic{figure}}
\renewcommand{\theHtable}{appendix.\arabic{table}}
\renewcommand{\theHequation}{appendix.\arabic{equation}}
\def\GraphSelectAppendixInput{}
This appendix extends the experiments, develops the theoretical results, and
describes the implementation and evaluation of \method{}. The additional
comparisons examine how selection changes with the available capacity,
predictor architecture, reference representation, and search procedure.

\section{Additional Experiments}
\label{sec:additional-results}

\subsection{Comparison Methods and Controls}

Table~\ref{tab:comparison-entries} summarizes the comparison methods and
controls. Published attribution, explanation, and sufficient-subset methods
provide alternative ways to score the candidate representations. The study
controls test the input used for scoring and the prediction objective, while
the method variants examine construction, exchange, and score updates.

\begin{table}[H]
\caption{\textbf{Selection methods and controls.} The rows summarize how each method selects representations.}
\label{tab:comparison-entries}
\centering
\graphselectCompactTableSetup
\begin{tabular}{@{}>{\raggedright\arraybackslash}p{0.30\linewidth}>{\raggedright\arraybackslash}p{0.64\linewidth}@{}}
\toprule
\tabhead{Entry} & \tabhead{Operation in this paper} \\
\midrule
\multicolumn{2}{l}{\textbf{Published method adapters}} \\
Shapley Sampling, Integrated Gradients, Feature Ablation, GNNExplainer, and GOAt & Published scoring or explanation methods adapted to rank the same candidate representations \\
Zorro & A published sufficient-subset method adapted to the same candidates and modality quotas \\
\midrule
\multicolumn{2}{l}{\textbf{Study controls}} \\
CE and TV deletion & Remove one candidate from the full input and measure cross entropy or total variation \\
KL divergence and prediction agreement & Build from the reference input by matching probabilities or predicted classes \\
\midrule
\multicolumn{2}{l}{\textbf{Method variants}} \\
Initial-gain ranking and Forward construction & Stop after the initial ranking or recompute exact addition gains after every selection \\
Exchange without score updates & Use the same initial set and evaluation cap as \mbox{\method{}} while keeping reference gains unchanged \\
\midrule
\multicolumn{2}{l}{\textbf{Proposed method}} \\
\method{} & Screen removals and additions at the current input and evaluate same-modality exchanges exactly \\
\bottomrule
\end{tabular}
\end{table}

\subsection{Deletion and Addition Alignment}

We compare deletion scores with exact first-addition gains for 6,600 candidate
representations from five graphs, three split seeds, and two modalities. CE
deletion identifies the best first addition in 19 of 30 graph, seed, and
modality groups, compared with 13 for TV deletion. Their mean Spearman
correlations with addition gains are 0.950 and 0.791. At the 5\% ranking depth,
the candidates selected by the two deletion rules recover 95.5\% and 85.4\%
of the average gain obtained by the exact addition ranking. Thus, deletion
scores retain useful information, but a candidate's value can change between
full input and the reference input.

In the construction study, Forward achieves lower loss than the initial-gain
ranking, KL divergence, prediction agreement, and both deletion controls in
all 54 graph, seed, and budget settings. Table~\ref{tab:design-controls}
reports the mean and standard deviation over the 18 graph and seed settings
at each budget. These are means of individual recovery ratios. The pooled
recovery in the main text is calculated after pooling the losses over affected
target nodes, so the two summaries can differ.

\begin{table}[H]
\caption{\textbf{Design controls.} Objective recovery is mean $\pm$ standard deviation.}
\label{tab:design-controls}
\centering
\small
\setlength{\tabcolsep}{5.5pt}
\renewcommand{\arraystretch}{1.04}
\begin{tabular}{lrrr}
\toprule
\tabhead{Control} & \tabhead{5\%} & \tabhead{10\%} & \tabhead{20\%} \\
\midrule
\multicolumn{4}{l}{\textbf{Input state}} \\
CE deletion & \tabstat{17.3}{3.9} & \tabstat{28.9}{6.0} & \tabsecondstat{47.0}{8.4} \\
TV deletion & \tabstat{14.9}{3.4} & \tabstat{25.1}{4.5} & \tabstat{42.0}{6.4} \\
\midrule
\multicolumn{4}{l}{\textbf{Prediction target}} \\
KL divergence & \tabstat{13.9}{3.2} & \tabstat{23.9}{5.0} & \tabstat{40.7}{6.6} \\
Prediction agreement & \tabstat{9.9}{2.8} & \tabstat{15.7}{4.2} & \tabstat{25.3}{4.7} \\
\midrule
\multicolumn{4}{l}{\textbf{Score updates}} \\
Initial-gain ranking & \tabsecondstat{17.5}{3.9} & \tabsecondstat{29.1}{5.9} & \tabstat{46.9}{8.2} \\
\greedy{} construction & \tabbeststat{17.8}{4.0} & \tabbeststat{29.8}{6.0} & \tabbeststat{48.0}{8.5} \\
\bottomrule
\end{tabular}
\end{table}

\subsection{Comparison with Full Input at Six Capacities}

The capacity study evaluates GraphSelect, Forward construction, the
initial-gain ranking, and CE deletion on Grocery and RedditS using nine trained
architectures, three seeds, 220 candidate nodes, and six quotas per modality.
For each setting, the selectors share the predictor, candidates, data split,
target nodes, and reference representations. We compare their predictions with
full input on the same test nodes, giving 1,296 paired measurements.

\begin{figure}[H]
\centering
\includegraphics[width=0.96\textwidth]{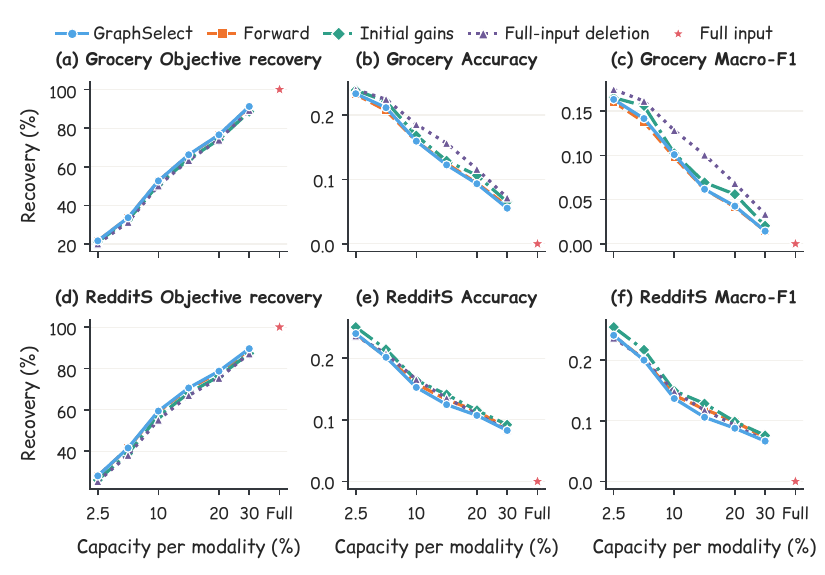}
\caption{\textbf{Selection across capacities.} Mean objective recovery and downstream performance gaps relative to full input for four selectors on Grocery and RedditS at six capacities.}
\label{fig:supp-selector-trajectories}
\end{figure}

GraphSelect has the highest mean objective recovery at every capacity
(Fig.~\ref*{fig:supp-selector-trajectories}). From 10\% onward, it also has the
smallest mean accuracy and Macro-F1 gaps. Forward is slightly closer to full
input on these two downstream metrics at 2.5\% and 5\%. The difference shows
that better recovery of predicted-class probability need not produce a closer
match in classification performance. Table~\ref{tab:budget-sweep-exact}
reports the means and standard deviations.

\begin{figure}[H]
\centering
\includegraphics[width=0.86\textwidth]{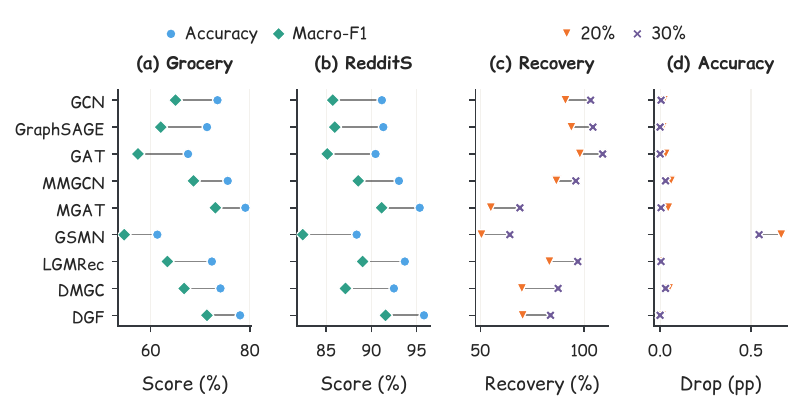}
\caption{\textbf{Variation across architectures.} Full-input prediction quality and changes in recovery and accuracy gap from 20\% to 30\% capacity for nine architectures.}
\label{fig:supp-architecture-context}
\end{figure}

The architectures differ in both full-input prediction quality and their
response to selection (Fig.~\ref*{fig:supp-architecture-context}). Increasing
capacity from 20\% to 30\% brings most architectures closer to their full-input
results. Recovery varies more than the accuracy gap, partly reflecting
differences in the reference-to-full-input loss reduction used to normalize
recovery. We report every architecture evaluated in this comparison.

Tables~\ref{tab:full-input-quality-nine} and
\ref{tab:architecture-budget-detail} report the corresponding means and
population standard deviations. MGAT has the highest full-input accuracy and
Macro-F1 on Grocery, whereas DGF leads both metrics on RedditS. GAT has the
highest recovery at 20\% and 30\%, but another architecture has the smallest
accuracy gap. The architecture with the largest recovery therefore need not
be the one with the closest downstream performance to full input.

\begin{table}[H]
\centering
\caption{\textbf{Full input quality.} Values summarize three seeds for nine architectures.}
\label{tab:full-input-quality-nine}
\graphselectCompactTableSetup
\begin{tabular}{@{}lrrrr@{}}
\toprule
\multirow{2}{*}{\tabhead{Architecture}} & \multicolumn{2}{c}{\tabhead{Grocery}} & \multicolumn{2}{c}{\tabhead{RedditS}} \\
\cmidrule(lr){2-3}\cmidrule(lr){4-5}
& \tabhead{Accuracy} & \tabhead{Macro-F1} & \tabhead{Accuracy} & \tabhead{Macro-F1} \\
\midrule
\multicolumn{5}{l}{\textbf{Graph architectures}} \\
GCN & \tabstat{73.47}{2.15} & \tabstat{65.01}{2.19} & \tabstat{91.16}{1.00} & \tabstat{85.73}{1.93} \\
GraphSAGE & \tabstat{71.36}{2.23} & \tabstat{62.03}{3.17} & \tabstat{91.32}{0.41} & \tabstat{85.96}{0.36} \\
GAT & \tabstat{67.54}{2.17} & \tabstat{57.43}{2.27} & \tabstat{90.46}{0.32} & \tabstat{85.12}{0.77} \\
\midrule
\multicolumn{5}{l}{\textbf{Multimodal graph architectures}} \\
MMGCN & \tabstat{75.54}{1.99} & \tabstat{68.62}{2.21} & \tabstat{93.05}{1.05} & \tabstat{88.56}{1.44} \\
MGAT & \tabbeststat{79.09}{0.30} & \tabbeststat{73.03}{0.42} & \tabsecondstat{95.34}{0.34} & \tabsecondstat{91.12}{0.88} \\
GSMN & \tabstat{61.34}{0.55} & \tabstat{54.65}{0.34} & \tabstat{88.37}{0.44} & \tabstat{82.42}{1.08} \\
LGMRec & \tabstat{72.35}{0.79} & \tabstat{63.37}{0.74} & \tabstat{93.70}{0.10} & \tabstat{89.03}{0.26} \\
DMGC & \tabstat{74.08}{1.15} & \tabstat{66.74}{1.49} & \tabstat{92.49}{0.48} & \tabstat{87.13}{0.62} \\
DGF & \tabsecondstat{78.03}{0.39} & \tabsecondstat{71.34}{0.98} & \tabbeststat{95.82}{0.33} & \tabbeststat{91.57}{1.04} \\
\bottomrule
\end{tabular}
\end{table}

\begin{table}[H]
\centering
\caption{\textbf{Architecture details.} Values summarize two graphs and three seeds.}
\label{tab:architecture-budget-detail}
\graphselectCompactTableSetup
\begin{tabular}{@{}lrrrr@{}}
\toprule
\multirow{2}{*}{\tabhead{Architecture}} & \multicolumn{2}{c}{\tabhead{20\% capacity}} & \multicolumn{2}{c}{\tabhead{30\% capacity}} \\
\cmidrule(lr){2-3}\cmidrule(lr){4-5}
& \tabhead{Recovery} & \tabhead{Accuracy drop} & \tabhead{Recovery} & \tabhead{Accuracy drop} \\
\midrule
\multicolumn{5}{l}{\textbf{Graph architectures}} \\
GCN & \tabstat{91.0}{4.7} & \tabstat{0.021}{0.034} & \tabstat{103.1}{5.8} & \tabstat{0.006}{0.028} \\
GraphSAGE & \tabsecondstat{93.8}{15.8} & \tabstat{0.015}{0.029} & \tabsecondstat{104.2}{14.4} & \tabbeststat{0.000}{0.018} \\
GAT & \tabbeststat{97.9}{9.4} & \tabstat{0.030}{0.043} & \tabbeststat{108.9}{10.7} & \tabbeststat{0.000}{0.046} \\
\midrule
\multicolumn{5}{l}{\textbf{Multimodal graph architectures}} \\
MMGCN & \tabstat{86.6}{7.4} & \tabstat{0.060}{0.061} & \tabstat{96.0}{4.1} & \tabstat{0.030}{0.038} \\
MGAT & \tabstat{55.1}{3.5} & \tabstat{0.044}{0.066} & \tabstat{69.1}{3.3} & \tabsecondstat{0.005}{0.028} \\
GSMN & \tabstat{50.5}{4.4} & \tabstat{0.667}{0.145} & \tabstat{64.2}{4.2} & \tabstat{0.545}{0.121} \\
LGMRec & \tabstat{83.2}{7.7} & \tabbeststat{0.006}{0.068} & \tabstat{96.9}{4.6} & \tabsecondstat{0.005}{0.033} \\
DMGC & \tabstat{70.0}{7.4} & \tabstat{0.050}{0.032} & \tabstat{87.5}{9.5} & \tabstat{0.030}{0.034} \\
DGF & \tabstat{70.4}{3.0} & \tabsecondstat{0.010}{0.040} & \tabstat{83.8}{3.8} & \tabbeststat{0.000}{0.034} \\
\bottomrule
\end{tabular}
\end{table}

\subsection{Reference and Search Controls}

\suppressfloats[t]
\begin{table}[t]
\caption{\textbf{Search ablations over 126 settings excluded from tuning.}}
\label{tab:search-ablation}
\centering
\graphselectSearchTableSetup
\begin{tabular}{@{}>{\raggedright\arraybackslash}p{0.25\linewidth}>{\raggedright\arraybackslash}p{0.34\linewidth}rr@{}}
\toprule
\tabhead{Variant} & \tabhead{Modification} & \shortstack[c]{$R_{\rm variant}-R_{\rm GraphSelect}$\\Loss $\times10^4$} & \shortstack[c]{GraphSelect\\W/T/L} \\
\midrule
\multicolumn{4}{l}{\textbf{Construction alternative}} \\
Forward construction & Repeated additions with rescoring & \tabstat{0.81}{3.75} & 57/68/1 \\
\midrule
\multicolumn{4}{l}{\textbf{Component removals}} \\
w/o exchange refinement & Return the initial set & \tabstat{4.60}{10.67} & 112/14/0 \\
w/o current-state updates & Keep initial scores during exchange & \tabstat{0.76}{3.85} & 62/62/2 \\
\bottomrule
\end{tabular}
\end{table}

We repeat the comparison with four reference rules. They are the training mean,
training median, nearest training representation to the mean, and zero vector.
GraphSelect achieves lower mean loss than the initial-gain ranking under every rule.
The mean Jaccard overlap with the training mean selection is 0.879 for the
median, 0.348 for the nearest representation, and 0.538 for zero. The method
ordering remains stable even though the selected sets vary across references.

The nonlinear search comparison uses 126 settings from the nine architectures,
excluding the settings used to tune the shortlist width. GraphSelect has W/T/L
counts of 57/68/1 against Forward construction and 112/14/0 against the version
without exchange refinement. It uses 43.60\% of the full forward passes
available under the matched evaluation budget. Exchange without score updates
starts from the same initial set and receives the same evaluation budget per
setting, but uses the initial gains throughout refinement. GraphSelect has
W/T/L counts of 62/62/2 against this control. Recomputing the scores improves 62 settings and preserves the loss within the
tie tolerance in another 62. Table~\ref{tab:search-ablation} gives the mean loss
differences and their variation across settings.

\begin{figure}[H]
\centering
\includegraphics[width=0.86\linewidth]{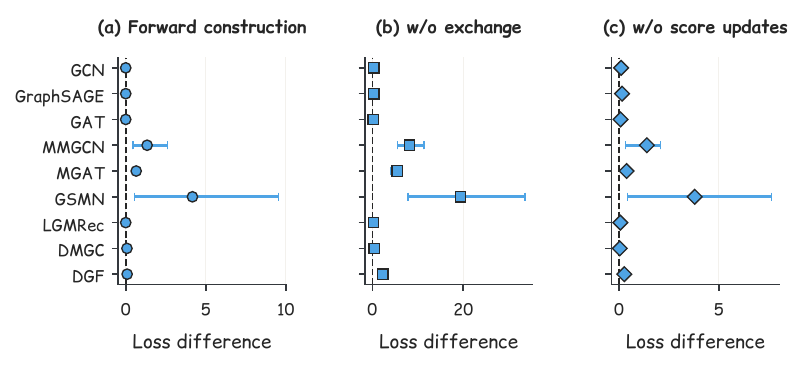}
\caption{\textbf{Search ablations across architectures.} Loss differences between GraphSelect and its search controls, scaled by $10^4$. Positive values favor GraphSelect.}
\label{fig:supp-search-transfer}
\end{figure}

Figure~\ref*{fig:supp-search-transfer} shows how these differences vary across
architectures. Score updates give the largest gains for MMGCN, MGAT, and GSMN,
while several other architectures remain near a tie. A batch exchange control
on the additive predictor provides a further comparison under the same
evaluation budget. It improves several individual settings at 5\% and 10\%
capacity. At 20\%, screened exchange is better in 13 settings, tied in 5, and
worse in none. Thus, the benefit of screening also depends on the capacity.

\FloatBarrier

\subsection{Algorithm Details}

The larger search uses $L\le\lceil\sqrt N\rceil$, giving at most 15 candidates per shortlist for $N=220$.

GraphSelect takes a trained predictor, full-input target classes, candidate
representations, modality references, quotas, shortlist width, objective
tolerance, and evaluation budget. It returns a subset that satisfies both
modality quotas. Initialization proceeds as follows.

\begin{enumerate}
\item Build the reference input and evaluate its loss.
\item Add each candidate alone and compute its exact gain.
\item Rank candidates within each modality using the predetermined tie order.
\item Select the first $K_m$ candidates from every modality.
\item Evaluate the loss of the initialized subset.
\end{enumerate}

The exchange stage repeats the following steps until it stops.

\begin{enumerate}
\item Evaluate the loss after removing each selected candidate.
\item Evaluate the loss after adding each unselected candidate.
\item Keep at most $L$ candidates from each ranking within each modality.
\item Evaluate every feasible exchange between the resulting shortlists.
\item Accept the best exchange only when its gain exceeds $\varepsilon$.
\item Recompute both rankings after an accepted exchange or stop if no pair improves the loss.
\end{enumerate}

Initialization, screening, and joint exchange evaluations all count toward the
evaluation budget. A deterministic candidate order resolves exact ties. After
each accepted exchange, the implementation evaluates the selected input again
to verify the incremental update.

\subsection{Exact Optima and Larger Candidate Pools}

The exact oracle comparison uses five predefined eight-node pools for every graph and seed.
At quotas one, two, and three per modality, Forward obtains 99.44\%, 99.77\%,
and 99.46\% of the reference-to-oracle improvement on average. Exact set matches
occur in 88.9\%, 91.1\%, and 74.4\% of the settings. These small pools allow
comparison with the exact optimum and show that Forward often finds a nearly
optimal subset. An exact optimum is not available for the larger $N=220$
candidate pools.

With 1,000 candidate nodes and quotas of 50, 100, and 200 per modality, Forward
achieves lower loss than the initial-gain ranking in all 54 settings. Reusing
gains whose affected target rows have not changed reduces the number of exact
marginal evaluations by 85.75\% on average. This reduction measures the work
avoided by sparse evaluation. End-to-end runtime was not measured in this test.

Direct recomputation of the reference logits, accepted exchanges, final subset,
and full-input predictions agrees with the incremental implementation within
the numerical tolerance. The nonlinear implementation uses a full forward pass
for each objective evaluation.

\section{Theory}
\label{sec:theory}

\subsection{Task and Search Rule}

Let $P_S$ be the predictor probabilities when selected representations use
their observed values and all other candidates use modality references. Let
$\widehat y_t$ be the class predicted from the complete candidate pool. The
selection loss is
\begin{equation}
R(S)=\frac{1}{|\cT|}\sum_{t\in\cT}-\log P_{S,t\widehat y_t}.
\label{eq:supp-risk}
\end{equation}
The constrained task keeps an exact quota for every modality.
\begin{equation}
S^\star\in\underset{S\subseteq\cE,\ |S\cap\cE_m|=K_m\ \forall m}{\arg\min}\ R(S).
\label{eq:supp-task}
\end{equation}
GraphSelect evaluates a same-modality exchange by jointly removing and adding a representation.
\begin{equation}
\Gamma(r,a\mid S)=R(S)-R\bigl((S\setminus\{r\})\cup\{a\}\bigr).
\label{eq:supp-exchange}
\end{equation}
The full-input predicted classes are determined before selection. Neither
ground-truth labels nor test membership enters $R$. A text representation can
be exchanged only for another text representation, and the same rule applies
to images.

\noindent\textbf{Proposition 1.}
Every subset accepted by GraphSelect satisfies the modality quotas. Every accepted move
strictly lowers $R$. The algorithm terminates under a finite objective
evaluation cap.

\noindent\textbf{Proof.}
The initial set contains exactly $K_m$ representations from every modality.
Each move removes and adds one representation from the same modality, so the
counts remain unchanged. The acceptance rule requires
$\Gamma(r,a\mid S)>\varepsilon$, which gives
$R(S')<R(S)$. Every screen and exchange consumes a positive integer number of
objective evaluations. A finite budget therefore permits only finitely many trials.
\hfill$\square$

The proposition establishes feasibility, loss reduction after each accepted
exchange, and termination. Because screening can omit an improving exchange,
it guarantees neither local nor global optimality.

\subsection{A Deletion Ranking Counterexample}

The direction of a transition can change a candidate ranking even without graph
propagation. Consider a smooth binary predictor. Let $\cE=\{a,b,c\}$, let $x_e\in[0,1]$, and choose
$0<\varepsilon<1$ and $0<\delta<1/2$. Define
\begin{equation}
\begin{aligned}
q(x)&=x_a+x_b-x_ax_b-\varepsilon x_ax_b(1-x_c),\\
p_x&=\bigl(\delta+(1-2\delta)q(x),\ 1-\delta-(1-2\delta)q(x)\bigr).
\end{aligned}
\label{eq:supp-counterexample}
\end{equation}
The full-input deletion score measures the total variation distance caused by removing a candidate.
\begin{equation}
s_{\mathrm{del}}(e)=\TV\bigl(p_{(1,1,1)},p_{(1,1,1)-\mathbf e_e}\bigr).
\label{eq:supp-deletion-score}
\end{equation}
The polynomial satisfies $0\le q(x)\le1$, so the probabilities are valid and
bounded away from zero. At the complete input, deleting $a$ or $b$ leaves
$q=1$, while deleting $c$ gives $q=1-\varepsilon$. The total variation rule therefore
chooses $c$. From the reference input, adding $c$ gives no loss reduction, while
adding $a$ or $b$ gives
\begin{equation}
\Delta(a\mid\varnothing)=\Delta(b\mid\varnothing)
=\log\frac{1-\delta}{\delta}>0.
\label{eq:supp-counterexample-gain}
\end{equation}
The deletion rule obtains zero addition gain while the optimum is positive.
Thus, this deletion ranking alone provides no positive constant-factor guarantee for addition gain.

\subsection{Finite Transition Identities}

For a sequence $S_0,S_1,\ldots,S_J$, write
$p_t^j=P_{S_j,t\widehat y_t}$. The finite gain of one transition is
\begin{equation}
\delta_j=R(S_{j-1})-R(S_j)
=\frac{1}{|\cT|}\sum_{t\in\cT}\log\frac{p_t^j}{p_t^{j-1}}.
\label{eq:supp-transition-gain}
\end{equation}
Summing the transition gains cancels every intermediate loss.
\begin{equation}
\sum_{j=1}^{J}\delta_j=R(S_0)-R(S_J).
\label{eq:supp-path}
\end{equation}
The loss is also the negative logarithm of the geometric mean probability assigned to the full-input predicted classes.
\begin{equation}
G(S)=\left(\prod_{t\in\cT}P_{S,t\widehat y_t}\right)^{1/|\cT|}
=\exp[-R(S)].
\label{eq:supp-geometric-mean}
\end{equation}
The identities apply to additions, removals, and exchanges.

For any threshold $\tau\in(0,1)$, define the low-confidence rate
$v_\tau(S)=|\cT|^{-1}\sum_t\1[p_t^S\le\tau]$. Let $d(S)$ be the disagreement
rate with the complete input classes. Then
\begin{equation}
v_\tau(S)\le\min\left\{1,\frac{R(S)}{-\log\tau}\right\}.
\label{eq:supp-confidence-bound}
\end{equation}
A class disagreement requires predicted class probability at most one half.
\begin{equation}
d(S)\le v_{1/2}(S)\le\min\left\{1,\frac{R(S)}{\log2}\right\}.
\label{eq:supp-disagreement-bound}
\end{equation}
The selected-input error is therefore bounded by the full-input error plus the disagreement rate.
\begin{equation}
\operatorname{Err}(S)\le\operatorname{Err}_{\mathrm{full}}+d(S).
\label{eq:supp-bounds}
\end{equation}
The first inequality follows because every term with $p_t^S\le\tau$ contributes
at least $-\log\tau$ to the loss. A class disagreement implies
$p_t^S\le1/2$, which gives the second line. The last line follows from the
pointwise union of full-input error and disagreement. These bounds relate the
selection loss to confidence and prediction error. They do not establish an
approximation ratio for subset selection.

\section{Exact Sparse Evaluation}
\label{sec:sparse-appendix}

For the additive predictor, each candidate induces a sparse change $B_e$ in
the logits of the target nodes. Its affected rows are
$\cT_e=\{t\in\cT\mid B_{e,t}\ne0\}$. The selected state is
\begin{equation}
Z_S=Z_\varnothing+\sum_{e\in S}B_e.
\label{eq:supp-sparse-state}
\end{equation}
Addition and exchange gains can then be evaluated using only the affected rows.
\begin{equation}
\begin{aligned}
\Delta(e\mid S)&=\frac{1}{|\cT|}\sum_{t\in\cT_e}
\bigl[\ell_t(Z_{S,t})-\ell_t(Z_{S,t}+B_{e,t})\bigr],\\
\Gamma(r,a\mid S)&=\frac{1}{|\cT|}\sum_{t\in\cT_r\cup\cT_a}
\bigl[\ell_t(Z_{S,t})-\ell_t(Z_{S,t}-B_{r,t}+B_{a,t})\bigr].
\end{aligned}
\label{eq:supp-sparse}
\end{equation}
Rows outside these sets cancel exactly. An inverted index maps target rows to
the candidates that affect them. An accepted update leaves a candidate's gain
unchanged when their affected rows are disjoint, allowing that gain to be
reused. We verify reused values by directly evaluating the objective.

Let $s_e=|\cT_e|$ and let $C$ be the class count. One candidate addition costs
$O(Cs_e)$ after the current logits are available. One exchange costs
$O(C|\cT_r\cup\cT_a|)$. These are arithmetic counts for the additive model.
For a nonlinear predictor, each objective evaluation requires a full forward
pass. We therefore distinguish objective evaluations, sparse row evaluations,
and full forward passes when reporting computational cost.

\subsection{Implementation and Computational Cost}

The sparse implementation caches three quantities. It stores each contribution
$B_e$ on its affected rows, the current logits, and candidate gains with their
dependent rows. After an accepted update, every gain that depends on a changed
row must be recomputed before reuse. This dependence determines which gains can be reused
without changing the result.

The nonlinear implementation constructs each candidate input, runs the trained
predictor, and evaluates the loss on the target nodes. Initialization and
screening count toward the budget of full forward passes. Loading the model
and encoding the features are common to all selectors and are excluded from
this count.

Both implementations follow the same selection rule. The additive structure
allows gains to be evaluated from sparse changes in logits, whereas the
nonlinear models require full forward passes. We evaluate selection quality
for both and report the corresponding computational cost separately.

\section{Experimental Protocol}
\label{sec:protocol}

\subsection{Datasets and Splits}

\begin{table}[t]
\caption{\textbf{Multimodal graph datasets.} Numbers of nodes, nonzero adjacency entries, and classes in the processed graphs used for evaluation.}
\label{tab:supp-datasets}
\centering
\graphselectCompactTableSetup
\begin{tabular}{lrrr}
\toprule
\multirow{2}{*}{\tabhead{Dataset}} & \multicolumn{2}{c}{\tabhead{Graph size}} & \multirow{2}{*}{\tabhead{Classes}} \\
\cmidrule(lr){2-3}
& \tabhead{Nodes} & \tabhead{Adjacency nonzeros} & \\
\midrule
SemArt & 21,382 & 1,173,014 & 10 \\
Grocery & 17,074 & 142,262 & 20 \\
Movies & 16,672 & 160,802 & 20 \\
Toys & 20,695 & 113,402 & 18 \\
RedditS & 15,894 & 283,080 & 20 \\
EleFashion & 97,766 & 399,172 & 11 \\
\bottomrule
\end{tabular}
\end{table}

We use the prescribed transductive graphs and three split seeds. Training
labels are used to fit the predictor, while training-node representations
define the modality references. Validation settings are used to tune the
search parameters. Test membership and labels are withheld during selection
and used only to evaluate the resulting predictions. Candidate nodes follow a
deterministic order. The main additive and nonlinear studies use the first
220 nodes, while the larger-scale comparison uses all 1,000 candidate nodes.

Each candidate node contributes one text representation and one image
representation. The 220-node setting therefore has 440 selectable units. The
5\%, 10\%, and 20\% budgets select 11, 22, and 44 units per modality. The full
capacity sweep adds quotas 6, 33, and 66 for 2.5\%, 15\%, and 30\%.

\subsection{Predictors and References}

The additive predictor combines modality logits, normalized one-hop and two-hop
propagation, and smoothed label propagation. Its parameters do not change
during selection. The nonlinear study uses GCN, GraphSAGE, GAT, MMGCN, MGAT, GSMN,
LGMRec, DMGC, and DGF
\citep{GCN,GraphSAGE,GAT,MMGCN,MGAT,GSMN,LGMRec,DMGC,DGF}.
Each architecture uses the stored text and image encodings. We evaluate all
trained models without retraining them or choosing checkpoints from the subset
results. Model evaluation is deterministic.

The main reference for each modality is the mean training representation.
The reference comparison also uses the training median, the nearest training
vector to the mean, and the zero vector. Each reference is determined before
candidate scoring and remains unchanged throughout selection.

\subsection{Selection and Evaluation Procedure}

For each graph and seed, selection follows model training and precedes test evaluation.

\begin{enumerate}
\item Fit or load the predictor using the prescribed training and validation data.
\item Set the trained predictor to deterministic evaluation mode.
\item Prepare the candidate order and modality references.
\item Evaluate the predictor on full candidate input to determine the target classes.
\item Run each selector without test membership or test labels.
\item Evaluate the selected inputs on test nodes.
\end{enumerate}

The complete candidate input is available before selection to determine the
target classes. This preparatory evaluation consumes no representation slots.
Selection then uses predicted-class probabilities without access to test
labels. It must be repeated if the predictor, candidate pool, reference,
target set, or budget changes.

We tune the nonlinear shortlist width on a predefined subset of Grocery
settings. The remaining Grocery settings and all RedditS settings provide a
disjoint evaluation of the selected width. The downstream comparison then
applies the same rule at all six capacities and compares the predictions with
full input.

\subsection{Comparators}

We adapt Shapley Sampling, Integrated Gradients, Feature Ablation,
GNNExplainer, GOAt, and Zorro to select whole representations. Each method
uses the same predictor, candidate pool, full-input target classes, and
separate modality quotas. Shapley Sampling uses a predefined sample budget.
Integrated Gradients follows the
straight path from the modality reference to the observed representation.
Feature Ablation replaces one complete representation group. GNNExplainer learns
a source-node mask and maps it to the two representation groups. GOAt
aggregates its attribution score over the target rows. Zorro uses a
predefined search width and fresh randomized complements to estimate its
class-retention score.

The study controls change one selection choice at a time. CE deletion and TV
deletion score a removal from full input. The initial-gain ranking scores each
candidate once at the reference input. KL divergence compares the full class
probability vector, while prediction agreement considers only whether the
predicted class changes. Forward construction recomputes exact addition gains
after every selected representation. Exchange without score updates and
GraphSelect start from the same initial set and receive the same budget of
full forward passes.

\subsection{Metrics and Statistical Units}

The primary objective is the cross-entropy $R(S)$ to the full-input predicted
classes. Objective recovery divides the loss reduction from the reference
input by the reduction obtained with full input. Recovery can exceed 100\%
when a subset assigns greater probability to those classes than full input
does, because $R(S)$ is not a distance between probability distributions.
Accuracy, Macro-F1, cross-entropy to ground-truth labels, prediction agreement,
and total variation describe other changes in the predictions. Comparisons
with full input use the same trained model and target nodes.

Tables report mean $\pm$ population standard deviation whenever repeated
settings form the displayed unit. Main additive tables use 18 graph and seed
settings at each budget. The nonlinear tables use 54 graph, architecture, and
seed settings at each budget. Hierarchical uncertainty intervals resample
graphs and then seeds. Paired W/T/L counts compare losses within each setting,
using the stated numerical tolerance to identify ties.

\subsection{Reproducibility Details}

The source package includes experiment configurations, commands, software
environment details, individual results, and aggregation scripts. The scripts
require one result for each expected setting and pair methods by predictor,
candidate order, data split, and evaluation set. Figures and tables are
derived from these paired results.

\subsection{Graph Ablation Study}
\label{sec:graph-ablation}

We separate the contribution of graph information to prediction from the effect
of propagating candidate representations. The study uses the six additive
datasets, three split seeds, 220 candidate nodes, and 5\%, 10\%, and 20\%
capacity per modality. We compare three predictor conditions. The full graph
condition retains all terms. The second condition removes propagation of each
candidate's deviation from its modality mean, retaining its within-node term
and the original reference logits. Noncandidate graph context and label
propagation remain available. The third condition removes both feature
propagation and label propagation, retaining the original weight on the
within-node modality logits.

Each condition defines its target classes using its own full candidate input.
The encodings, probe training procedure, splits, and candidate sets are shared
across conditions. The modality probes are reconstructed from the same recipe
and seeds. This reconstruction changes no full-input predicted classes relative
to the original saved probabilities. The within-node predictor uses the original
parameters without additional tuning. These controlled ablations examine the
additive model evaluated here.

% Derived from the graph-dependence table calculation.
\begin{table}[t]
\centering
\small
\caption{Full-input test accuracy in the graph ablation study. Entries are mean $\pm$ population standard deviation over three seeds, in percent. The last column gives the mean number of target nodes affected by one candidate representation in the full graph predictor. Both ablations reduce this number to one.}
\label{tab:graph-ablation-accuracy}
\begin{tabular}{lrrrr}
\toprule
Dataset & Full graph & \shortstack{No candidate\\propagation} & No graph terms & \shortstack{Affected\\targets} \\
\midrule
SemArt & $84.11 \pm 0.09$ & $84.09 \pm 0.11$ & $83.42 \pm 0.91$ & 277.3 \\
Grocery & $75.01 \pm 0.61$ & $75.01 \pm 0.59$ & $60.56 \pm 5.47$ & 183.9 \\
Movies & $47.55 \pm 1.26$ & $47.56 \pm 1.24$ & $38.33 \pm 0.43$ & 108.7 \\
Toys & $76.85 \pm 0.64$ & $76.86 \pm 0.61$ & $67.42 \pm 0.97$ & 55.4 \\
RedditS & $94.69 \pm 0.30$ & $94.67 \pm 0.28$ & $90.53 \pm 0.36$ & 17.4 \\
EleFashion & $80.06 \pm 0.38$ & $80.06 \pm 0.38$ & $82.21 \pm 0.25$ & 120.6 \\
\midrule
Mean & 76.38 & 76.38 & 70.41 & 127.2 \\
\bottomrule
\end{tabular}
\end{table}

\begin{table}[t]
\centering
\small
\caption{Selection losses in the graph ablation study. Each entry averages 18 graph and seed settings. The first two losses use each condition\textquotesingle s own full-input target classes. The two rightmost columns report the control loss minus the \method{} loss, multiplied by $10^5$. Positive values favor \method{}.}
\label{tab:graph-ablation-selection}
\setlength{\tabcolsep}{3pt}
\begin{tabular}{llrrrr}
\toprule
Predictor & Capacity & $R(\varnothing)$ & $R(\cE)$ & \shortstack{Initial ranking\\minus \method{}} & \shortstack{\greedy{}\\minus \method{}} \\
\midrule
Full graph & 5\% & 0.522555 & 0.516625 & 1.599 & 0.352 \\
 & 10\% & 0.522555 & 0.516625 & 4.026 & 0.776 \\
 & 20\% & 0.522555 & 0.516625 & 6.567 & 1.756 \\
\midrule
No candidate propagation & 5\% & 0.522529 & 0.517774 & 2.317 & 0.557 \\
 & 10\% & 0.522529 & 0.517774 & 3.728 & 0.983 \\
 & 20\% & 0.522529 & 0.517774 & 5.499 & 1.254 \\
\midrule
No graph terms & 5\% & 1.555942 & 1.544582 & 1.185 & 0.268 \\
 & 10\% & 1.555942 & 1.544582 & 3.055 & 0.693 \\
 & 20\% & 1.555942 & 1.544582 & 6.140 & 2.535 \\
\bottomrule
\end{tabular}
\end{table}

Table~\ref{tab:graph-ablation-accuracy} shows that graph terms improve mean
full-input test accuracy by 5.97 percentage points. They improve accuracy on
five datasets. On EleFashion, removing graph terms instead raises accuracy
from 80.06\% to 82.21\%. Removing only candidate propagation leaves most of the
graph context available, and the corresponding mean accuracy changes by less
than 0.01 percentage points. Nevertheless, propagation changes the reach of
each candidate representation. Its mean number of affected targets ranges
from 17.4 on RedditS to 277.3 on SemArt, compared with one in either ablation.
Different candidate nodes share affected targets in 12.1\% of pairs on average
with propagation and in none of the pairs after it is removed.

For selection, \greedy{} repeatedly evaluates all feasible additions until the
quotas are filled. \method{} uses a shortlist of at most eight candidates per
modality and 35\% of the analytical upper bound on these \greedy{} evaluations.
The initial-gain ranking evaluates candidates once. Each method keeps the same
evaluation-limit rule across all three predictor conditions. This diagnostic
uses a common analytical limit to compare conditions, while the main selection
tables use their original matched \greedy{} evaluation counts.

Table~\ref{tab:graph-ablation-selection} reports the reference and full-input
losses alongside raw selection-loss improvements. These losses are averaged
over all target nodes. Recovery ratios have different denominators and target
classes across conditions. Within the full graph condition, \method{} improves
on the initial-gain ranking in all 54 settings and on \greedy{} in 47, with six
ties and one loss. Without candidate propagation, the corresponding W/T/L
counts are 53/1/0 and 47/6/1. Without either graph term, they are 51/3/0 and
45/8/1. Ties use a tolerance of $10^{-12}$. Exchanges therefore improve input
combinations in all three conditions. Graph propagation determines their
cross-node effects, while modality fusion and the prediction loss also make a
representation's gain depend on the selected subset.

\textbf{Illustrative exchange.}
Figure~\ref{fig:task-setting} uses Grocery seed 44 at 20\% capacity.
The target is node 13190, which is outside the candidate pool. The neighboring
jicama is node 13931, connected by the recorded also-viewed relation. The fern is node
8249, outside the target's two-hop neighborhood. The third accepted exchange
replaces the fern image with the neighboring jicama image in the selected set, preserving
44 text and 44 image representations. Direct replay gives target-class
probabilities of 47.5964\% before and 53.0059\% after, compared with 53.4605\%
under full input. The predicted category remains Produce throughout. Mean
selection loss over all targets falls from 0.7554431450 to 0.7554343851.
At the reference input, the fern and neighboring jicama images have addition gains of
$5.0598\times10^{-5}$ and $4.3646\times10^{-5}$, respectively. Immediately before
the exchange, removing the fern image increases loss by $3.4887\times10^{-5}$,
while adding the neighboring jicama image reduces it by $4.3646\times10^{-5}$.
The figure shows the original product photographs and unchanged product
identities. The neighbor's text uses its modality mean in both limited-input states,
while the fern's text and both target inputs remain observed.

\subsection{Scope and Limitations}

Objective recovery measures improvement in the selection loss, while accuracy
and Macro-F1 measure classification performance on test nodes. The architecture
results show why both are needed. A high recovery ratio need not imply the
smallest accuracy gap, and an aggregate mean can conceal this variation.

The evaluated workflow uses complete candidate representations during selection
and a capacity-limited subset during subsequent inference. Each selection is
associated with a trained predictor, candidate pool, reference, target set, and
budget. A process that needs only previously computed predictions can cache those
outputs directly. Extending the analysis to storage bytes, latency, or energy
requires weighted costs and system measurements. Reconstruction of missing
modalities and transfer to future models involve additional modeling choices.

The graph ablation study examines the additive predictor. Extending it to the
nine nonlinear architectures would establish how architecture changes the
relationship between graph propagation and selection quality.

The theory guarantees quota feasibility, lower loss after an accepted exchange,
and termination under a finite evaluation budget. Screening can still miss an
improving exchange. The downstream experiments measure gaps to full input and
do not establish statistical equivalence. They support selection quality for
the studied workflow, while global optimality and deployment savings require
further investigation.

\FloatBarrier

\end{document}